\documentclass[10pt,conference]{IEEEtran}
\IEEEoverridecommandlockouts

\usepackage{dsfont}
\usepackage{algorithm}
\usepackage{CJKutf8}
\usepackage{balance}

\usepackage{url}
\usepackage{breakurl}
\usepackage[breaklinks]{hyperref}

\usepackage[noend]{algpseudocode}
\newcommand{\implement}{\textit{Purity}}

\usepackage{cite}
\usepackage{amsmath,amssymb,amsfonts}
\usepackage{graphicx}
\usepackage{xcolor}
\def\BibTeX{{\rm B\kern-.05em{\sc i\kern-.025em b}\kern-.08em
    T\kern-.1667em\lower.7ex\hbox{E}\kern-.125emX}}
    
\begin{document}

\title{Testing Machine Translation via Referential Transparency}

\author{\IEEEauthorblockN{Pinjia He}
\IEEEauthorblockA{\textit{Department of Computer Science} \\
\textit{ETH Zurich}\\
Switzerland \\
pinjia.he@inf.ethz.ch}
\and
\IEEEauthorblockN{Clara Meister}
\IEEEauthorblockA{\textit{Department of Computer Science} \\
\textit{ETH Zurich}\\
Switzerland \\
clara.meister@inf.ethz.ch}
\and
\IEEEauthorblockN{Zhendong Su}
\IEEEauthorblockA{\textit{Department of Computer Science} \\
\textit{ETH Zurich}\\
Switzerland \\
zhendong.su@inf.ethz.ch}
}

\maketitle

\begin{abstract}
Machine translation software has seen rapid progress in recent years due to the advancement of deep neural networks. People routinely use machine translation software in their daily lives for tasks such as ordering food in a foreign restaurant, receiving medical diagnosis and treatment from foreign doctors, and reading international political news online. However, due to the complexity and intractability of the underlying neural networks, modern machine translation software is still far from robust and can produce poor or incorrect translations; this  can lead to misunderstanding, financial loss, threats to personal safety and health, and political conflicts. To address this problem, we introduce \textit{referentially transparent inputs (RTIs)}, a simple, widely applicable methodology for validating machine translation software. A referentially transparent input is a piece of text that should have similar translations when used in different contexts. Our practical implementation, {\implement}, detects when this property is broken by a translation. To evaluate RTI, we use {\implement} to test Google Translate and Bing Microsoft Translator with 200 unlabeled sentences, which detected 123 and 142 erroneous translations with high precision (79.3\% and 78.3\%). The translation errors are diverse, including examples of under-translation, over-translation, word/phrase mistranslation, incorrect modification, and unclear logic.
\end{abstract}


\section{Introduction}\label{sec:intro}
Machine translation software aims to fully automate translating text from a source language into a target language. In recent years, the performance of machine translation software has improved significantly largely due to the development of neural machine translation (NMT) models ~\cite{Zhang18ACL, Gehring17ACL, Vaswani17NeurIPS}. In particular, machine translation software (\textit{e.g.}, Google Translate~\cite{Wu16Arxiv} and Bing Microsoft Translator~\cite{Hassan18Arxiv}) is approaching human-level performance in terms of human evaluation. Consequently, more and more people are employing machine translation in their daily lives, for tasks such as reading news and textbooks in foreign languages, communicating while traveling abroad, and conducting international trade. This is reflected in the increased use of machine translation software: in 2016, Google Translate attracted more than 500 million users and translated more than 100 billion words per day~\cite{userdata}; NMT models have been embedded in various software applications, such as Facebook~\cite{tranFacebook} and Twitter~\cite{tranTwitter}.

Similar to traditional software (\textit{e.g.}, a Web server), machine translation software's reliability is of great importance. Yet, modern translation software has been shown to return erroneous translations, leading to misunderstanding, financial loss, threats to personal safety and health, and political conflicts~\cite{translation1, translation2, translation3, translation4, translation5, translation6}. 
This behavior can be attributed to the brittleness of neural network-based systems, which is exemplified in autonomous car software~\cite{Pei17SOSP, Tian18ICSE}, sentiment analysis tools~\cite{Alzantot18EMNLP, Iyyer18NAACL, Li19NDSS}, and speech recognition services~\cite{Carlini16Security, Qin19ICML}. Likewise, NMT models can be fooled by adversarial examples (\textit{e.g.}, perturbing characters in the source text~\cite{ Ebrahimi18COLING}) or natural noise (\textit{e.g.}, typos~\cite{Belinkov18ICLR}). The inputs generated by these approaches are mostly illegal, that is, they contain lexical (\textit{e.g.}, ``bo0k'') or syntactic errors (\textit{e.g.}, ``he home went''). However, inputs to machine translation software are generally lexically and syntactically correct. For example, Tencent, the company developing WeChat, a messaging app with more than one billion monthly active users, reported that its embedded NMT model can return erroneous translations even when the input is free of lexical and syntax errors~\cite{ Zheng18Arxiv}. 

There remains a dearth of automated testing solutions for machine translation software---at least in part because the problem is quite challenging. First, most of the existing parallel corpora that could be used for testing have already been employed in the model training process. Thus, testing oracles of high quality are lacking. Second, in contrast to traditional software, the logic of neural machine translation software is largely embedded in the structure and parameters of the underlying model. Thus, existing code-based testing techniques cannot directly be applied to testing NMT. Third, existing testing approaches for AI (artificial intelligence) software~\cite{Pei17SOSP, Goodfellow15ICLR, Alzantot18EMNLP, Iyyer18NAACL, Li19NDSS} mainly target much simpler use cases (\textit{e.g.}, 10-class classification) and/or with clear oracles~\cite{Mudrakarta18ACL, Jia17EMNLP}. In contrast, testing the correctness of translations is a more complex task: source text could have multiple correct translations and the output space is magnitudes larger. Last but not least, existing machine translation testing techniques~\cite{He20ICSE, Sun20ICSE} generate test cases (\textit{i.e.}, synthesized sentences) by replacing one word in a sentence via language models. Thus, their performance is limited by the proficiency of existing language models.

\begin{figure*}[t]
\centering{} 
\includegraphics[scale=0.95]{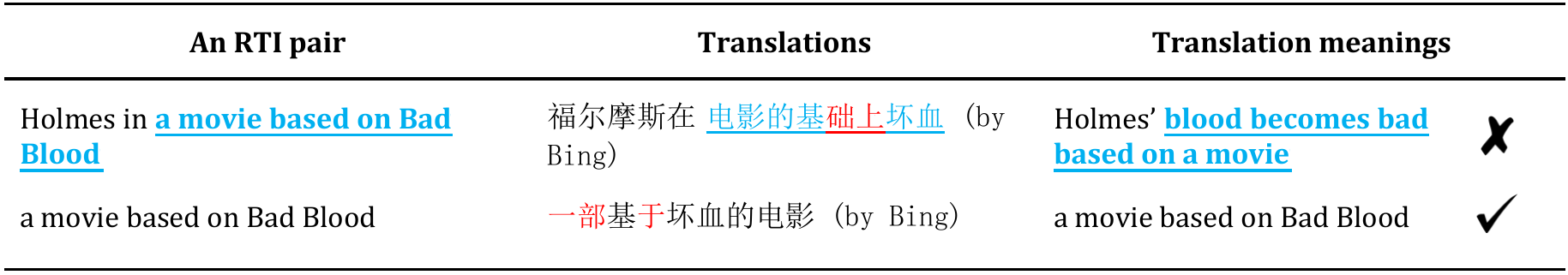}
\caption{Example of a referentially transparent input pair. The underlined phrase in the left column is an RTI extracted from the sentence. The differences in the translations are highlighted in red and their meanings are given in the right column. This RTI pair and its translations were reported by our approach as a suspicious issue. The first translation is erroneous.}
\label{fig:insight}
\end{figure*}

We introduce \textit{RTIs (referentially transparent inputs)}, a novel and general concept, as a method for validating machine translation software. The core idea of RTI is inspired by \textit{referential transparency}~\cite{Sondergaard90RefTrans, referentialtransparency}, a concept in programming languages (specifically functional programming): a method should always return the same value for a given argument. In this paper, we define a \textit{referentially transparent input (RTI)} as a piece of text that should have similar translations in different contexts. For example, ``a movie based on Bad Blood" in Fig.~\ref{fig:insight} is an RTI. The key insight is to generate a pair of texts that contain the same RTI and check whether its translations in the pair are similar. To realize this concept, we implement {\implement}, a tool that extracts phrases from arbitrary text as RTIs. Specifically, given unlabeled text in a source language, {\implement} extracts phrases via a constituency parser~\cite{SRConstParsing} and constructs \textit{RTI pairs} by grouping an RTI with either its containing sentence or a containing phrase. If a large difference exists between the translations of the same RTI, we report this pair of texts and their translations as a suspicious issue. The key idea of this paper is conceptually different from existing approaches~\cite{He20ICSE, Sun20ICSE}, which replace a word (\textit{i.e.}, the context is fixed) and assume that the translation should have only small changes. 
In contrast, this paper assumes that the translation of an RTI should be similar across different sentences/phrases (\textit{i.e.}, the context is varied).

We apply {\implement} to test Google Translate~\cite{googletranslate}  and Bing Microsoft Translator~\cite{bingmicrosofttranslator} with 200 sentences crawled from CNN by He \textit{et al}.~\cite{He20ICSE}. {\implement} successfully reports 154 erroneous translation pairs in Google Translate and 177 erroneous translation pairs in Bing Microsoft Translator with high precision (79.3\% and 78.3\%), revealing 123 and 142 erroneous translations respectively.\footnote{One erroneous translation could appear in multiple erroneous translation pairs (i.e., erroneous issues).} The translation errors found are diverse, including under-translation, over-translation, word/phrase mistranslation, incorrect modification, and unclear logic. Compared with the state-of-the-art~\cite{He20ICSE, Sun20ICSE}, {\implement} can report more erroneous translations with higher precision. Due to its conceptual difference, {\implement} can reveal many erroneous translations that have not been found by existing approaches (illustrated in Fig.~\ref{fig:venn}). Additionally, {\implement} spent 12.74s and 73.14s on average for Google Translate and Bing Microsoft Translator respectively, achieving comparable efficiency to the state-of-the-art methods. RTI's source code and all the erroneous translations found are released~\cite{bugsfound} for independent validation. The source code will also be released for reuse. The main contributions of this paper are as follows:
\begin{itemize}
    \item The introduction of a novel, widely-applicable concept, \textit{referentially transparent input (RTI)}, for systematic machine translation validation, 
    \item A realization of RTI, \implement, that adopts a constituency parser to extract phrases and a bag-of-words (BoW) model to represent translations, and
    \item Empirical results demonstrating the effectiveness of RTI: based on 200 unlabeled sentences, {\implement} successfully found 123 erroneous translations in Google Translate and 142 erroneous translations in Bing Microsoft Translator with 79.3\% and 78.3\% precision, respectively.
\end{itemize}

\section{Preliminaries}\label{sec:motivation}

\subsection{Referential Transparency}
In the programming language field, referential transparency refers to the ability of an expression to be replaced by its corresponding value in a program without changing the result of the program~\cite{Sondergaard90RefTrans, referentialtransparency}. For example, mathematical functions (\textit{e.g.}, square root function) are referentially transparent, while a function that prints a timestamp is not. 

Referential transparency has been adopted as a key feature by functional programming because it allows the compiler to reason about program behavior easily, which further facilitates higher-order functions (\textit{i.e.}, a series of functions can be glued together) and lazy evaluation (\textit{i.e.}, delay the evaluation of an expression until its value is needed)~\cite{RTadvantages}. The terminology ``referential transparency'' is used in a variety of fields with different meanings, such as logic, linguistics, mathematics, and philosophy. Inspired by the referential transparency concept in functional programming, a metamorphic relation can be defined within an RTI pair. 

\subsection{Metamorphic Relation}
\begin{figure*}[t]
\centering{}
 \includegraphics[scale=0.54]{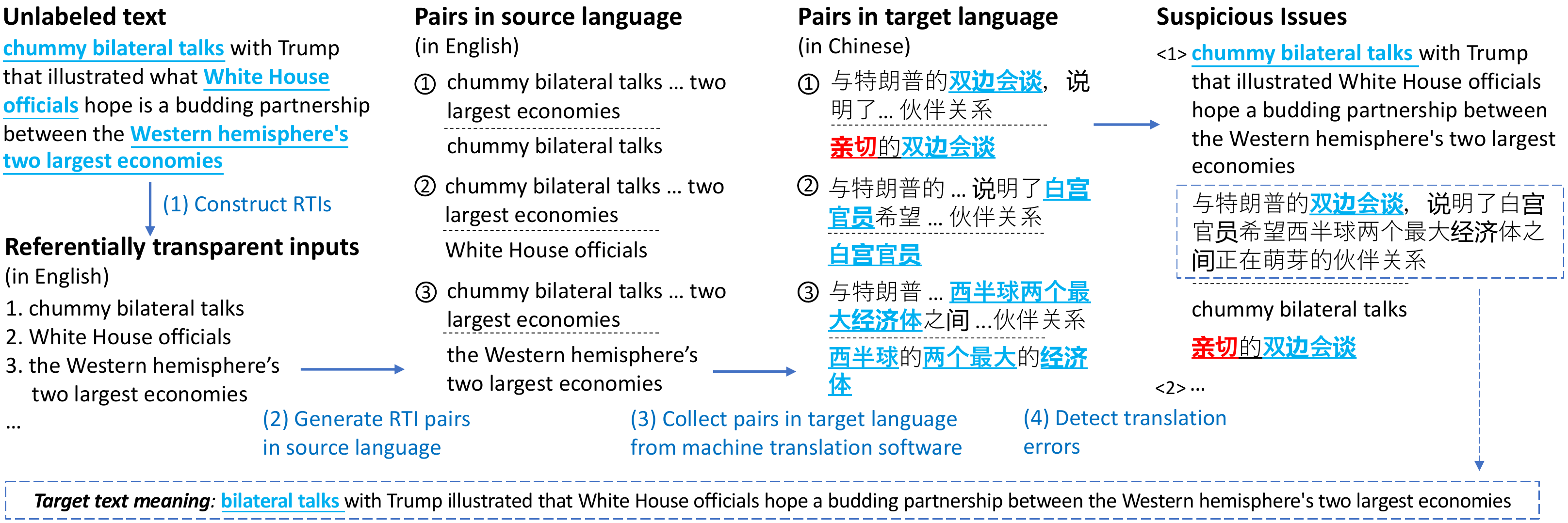} 
\caption{Overview of our RTI implementation. We use one English phrase as input for clarity and simplicity. In the ``Pairs in source language" column, the phrases above the dashed lines are the original unlabeled text. The texts marked in blue and underlined are RTIs or the identical characters in the translations of an RTI pair, while the texts marked in red are the characters in the translation of an RTI but not in that of its containing phrase/sentence.}
\label{fig:overview}
\end{figure*}

Metamorphic relations are necessary properties of functionalities of the software under test. In metamorphic testing~\cite{Chen98Metamorphic, Segura16TSE, Chen18CSUR}, the violation of a metamorphic relation will be suspicious and indicates a potential bug. We develop a metamorphic relation for machine translation software as follows: RTIs (e.g., noun phrases) should have similar translations in different contexts. Formally, for an RTI $r$, assume we have two different contexts $C_1$ and $C_2$ (i.e., different pieces of surrounding words). $C_1 (r)$ and $C_2 (r)$, which form an RTI pair, are the pieces of text containing $r$ and the two contexts respectively. To test the translation software $T$, we could obtain their translations $T(C_1 (r))$ and $T(C_2 (r))$. The metamorphic relation is defined as:
\begin{equation}
\label{equ:meta}
dist_r (T(C_1 (r)), T(C_1 (r))) \leq d,
\end{equation}
where $dist_r$ denotes the distance between the translations of $r$ in $T(C_1 (r))$ and in $T(C_2 (r))$; $d$ is a threshold controlled by the developers. In the following section, we will introduce our approach in detail with an example (Fig.~\ref{fig:overview}).

\section{RTI and {\implement}'s implementation}\label{sec:method}

This section introduces \textit{referentially transparent inputs (RTIs)} and our implementation, {\implement}. An RTI is defined as a piece of text that has similar translations across texts (\textit{e.g.}, sentences and phrases). Given a sentence, our approach intends to find its RTIs---phrases in the sentence that exhibit referential transparency---and utilize them to construct test inputs. To realize RTI's concept, we implement a tool called {\implement}.
The input of {\implement} is a list of unlabeled, monolingual sentences, while its output is a list of suspicious issues. Each issue contains two pairs of text: a base phrase (\textit{i.e.}, an RTI) and its container phrase/sentence, and their translations. Note that {\implement} should detect errors in the translation of either the base or container text. Fig.~\ref{fig:overview} illustrates the process used by {\implement}, which has the following four steps:

\begin{enumerate}
    \item \textit{Identifying referentially transparent inputs.} For each sentence, we extract a list of phrases as its RTIs by analyzing the sentence constituents.
    \item \textit{Generating pairs in source language.} We pair each phrase with either a containing phrase or the original sentence to form RTI pairs.
    \item \textit{Collecting pairs in target language.} We feed the RTI pairs to the machine translation software under test and collect their corresponding translations.
    \item \textit{Detecting translation errors.} In each pair, the translations of the RTI pair are compared with each other. If there is a large difference between the translations of the RTI, {\implement} reports the pair as potentially containing translation error(s).
\end{enumerate}
Algorithm~\ref{alg:rti} shows the pseudo-code of our RTI implementation, which will be explained in detail in the following sections.

\begin{algorithm}[h]
\small
\caption{RTI implemented as {\implement}.}
\label{alg:rti}
\begin{flushleft}
\textbf{Require:} $source\_sents$: a list of sentences in source language \\
\hspace{9ex} $d$: the distance threshold \\
\textbf{Ensure:} $suspicious\_issues$: a list of suspicious pairs
\end{flushleft}
\begin{algorithmic}[1]
\State $suspicious\_issues\leftarrow List(\hspace{0.5ex})$
\Comment{Initialize with empty list}
\ForAll{\textit{source\_sent} in \textit{source\_sents}}
	\State  $constituency\_tree\leftarrow$ \Call{parse}{\textit{source\_sent}}
	\State $head \gets constituency\_tree.$head(\hspace{0.5ex})
	\State $RTI\_source\_pairs\leftarrow List(\hspace{0.5ex})$ \State \Call{recursiveNPFinder}{$head$, List(\hspace{0.5ex}), $RTI\_source\_pairs$}
	\State $RTI\_target\_pairs\leftarrow$ \Call{translate}{$RTI\_source\_pairs$}
	\ForAll{ \textit{target\_pair} in \textit{RTI\_target\_pairs}}
	    \If{\Call{distance}{$target\_pair$} $> d$}
	         \State Add $source\_pair$, $target\_pair$ to $suspicious\_issues$
	    \EndIf
	\EndFor
\EndFor
	
\State \Return $suspicious\_issues$

\Statex 
\Function{recursiveNPFinder}{$node$,\hspace{0.5ex}$rtis$, $all\_pairs$}
    \If{$node$ is $leaf$}
        \State \Return
    \EndIf
    \If{$node$.constituent is NP}
        \State $phrase\leftarrow node$.string
        \ForAll{$container\_phrase$ in $rtis$}
            \State Add $container\_phrase$, $phrase$ to  $all\_pairs$
        \EndFor
        \State Add $phrase$ to  $rtis$
    \EndIf
    \ForAll{$child$ in $node$.children(\hspace{0.5ex})}
        \State \Call{recursiveNPFinder}{$child$, $rtis.\mathrm{copy}(\hspace{0.5ex})$, $all\_pairs$}
    \EndFor
\State \Return $all\_pairs$
\EndFunction

\Statex 
\Function{distance}{$target\_pair$}
	\State $rti\_BOW\gets$  \Call{bagOfWords}{$target\_pair$[0]}
	\State $container\_BOW\gets$  \Call{bagOfWords}{$target\_pair$[1]}
    \State \Return $|rti\_BOW \setminus container\_BOW|$
    
\EndFunction
\end{algorithmic}
\end{algorithm}

\subsection{Identifying RTIs}\label{sec:rtiidentify}
In order to collect a list of RTIs, we must find pieces of text with unique meaning, \textit{i.e.} their meaning should hold across contexts. To guarantee the lexical and syntactic correctness of RTIs, we extract them from published text (\textit{e.g.}, sentences in Web articles).

Specifically, {\implement} extracts noun phrases from a set of sentences in a source language as RTIs. For example, in Fig.~\ref{fig:overview}, the phrase ``chummy bilateral talks'' will be extracted; this phrase should have similar translations when used in different sentences (\textit{e.g.}, ``I attended chummy bilateral talks.'' and ``She held chummy bilateral talks.'') For the sake of simplicity and to avoid grammatically strange phrases, we only consider noun phrases in this paper.

\begin{figure}[t]
\centering{}
 \includegraphics[scale=0.65]{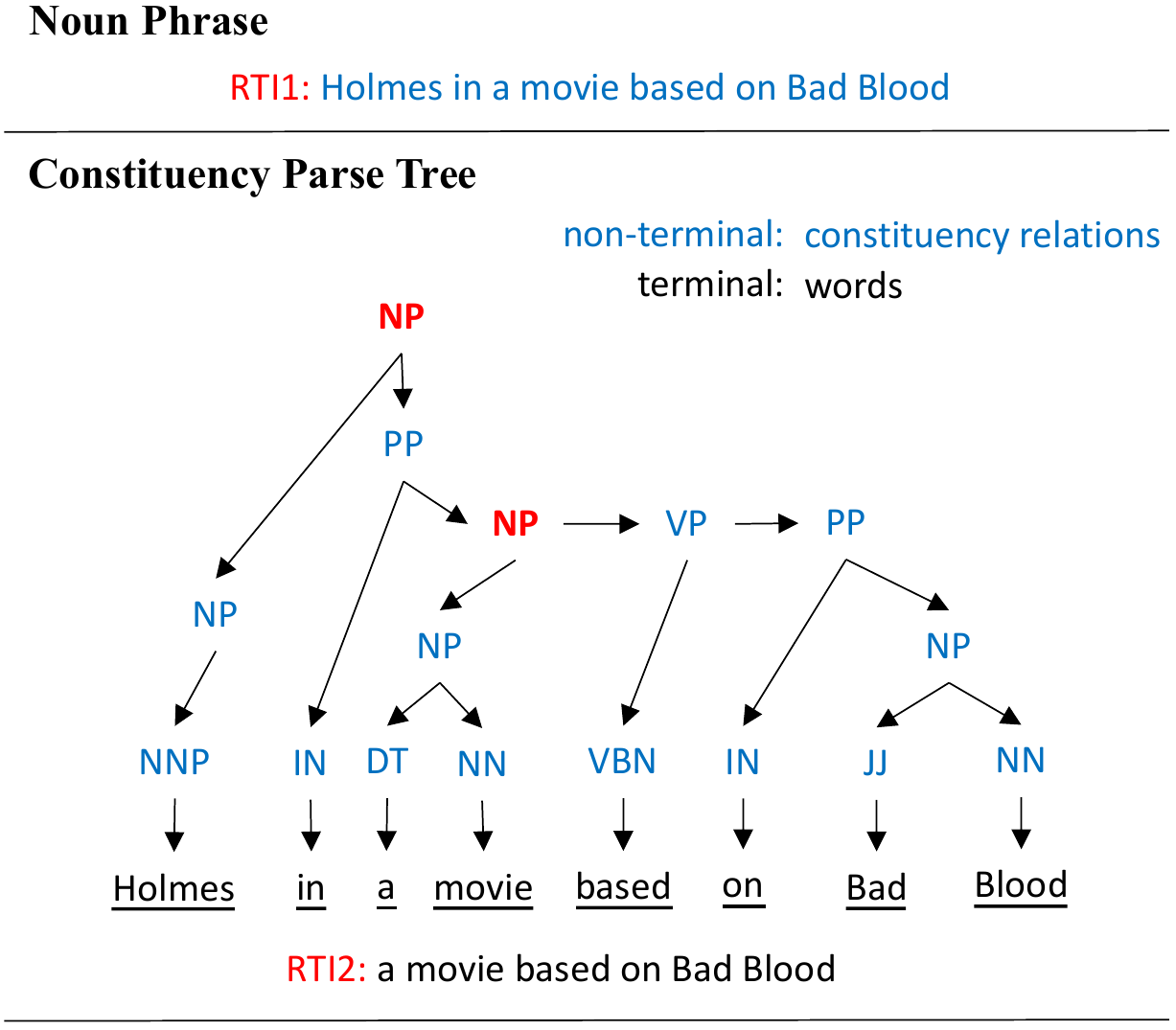} 
\caption{A constituency parse tree example. The non-terminal nodes in bold and red are the RTIs extracted by our approach.}
\label{fig:parsetree}
\end{figure}

We identify noun phrases using a constituency parser, a readily available natural language processing (NLP) tool. A constituency parser identifies the syntactic structure of a piece of text, outputting a tree where the non-terminal nodes are constituency relations and the terminal nodes are words (example shown in Fig.~\ref{fig:parsetree}). To extract all noun phrases, we traverse the constituency parse tree and pull out all the \textit{NP} (noun phrase) relations.

Note that in general, an RTI can contain another shorter RTI. For example, the second RTI pair in Fig.~\ref{fig:insight} contains two RTIs: ``Holmes in a movie based on Bad Blood'' is the containing RTI to ``a movie based on Bad Blood'' This holds true when noun phrases are used as RTIs as well, since noun phrases can contain other noun phrases.

Once we have obtained all the noun phrases from a sentence, we  filter out those containing more than 10 words and those containing less than 3 words\footnote{These filter values were tuned empirically via grid search on one dataset. In particular, we search the most suitable values in [1,20] and [2,10] with step size one for the two filters respectively. By most suitable values, we mean the filter values that achieve the highest ratio between the number of RTIs and the number of noun phrases after filtering.} that are not stop-words.\footnote{A stop-word is a word that is mostly used for structure---rather than meaning---in a sentence, such as ``is'', ``this'', ``an.''} This filtering helps us concentrate on unique phrases that are more likely to carry a single meaning and greatly reduces false positives. The remaining noun phrases are regarded as RTIs in {\implement}. 

\subsection{Generating Pairs in Source Language}
Once a list of RTIs has been generated, each must be paired with containing phrases, which will be used for referential transparency validation (Section~\ref{sec:detecterrors}). Specifically, each RTI pair should have two different pieces of text that contain the same phrase. To generate these pairs, we pair an RTI with the full text in which it was found (as in Fig.~\ref{fig:overview}) and with all the containing RTIs (\textit{i.e.}, other containing noun phrases) from the same sentence. For example, assume that RTI1 in Fig.~\ref{fig:parsetree} is an RTI extracted from a sentence, RTI2 can be found based on the constituency structure; note that ``Holmes in a movie based on Bad Blood" is the containing RTI to ``a movie based on Bad Blood". Thus, 3 RTI pairs will be constructed: (1) RTI1 and the original sentence; (2) RTI2 and the original sentence; and (3) RTI1 and RTI2.

\subsection{Collecting Pairs in Target Language}
The next step is to input RTI pairs (in the given source language) to the machine translation software under test and collect their translations (in any chosen target language). We use the APIs provided by Google and Bing in our implementation, which return results identical to Google Translate and Bing Microsoft Translator's Web interfaces~\cite{googletranslate, bingmicrosofttranslator}.

\subsection{Detecting Translation Errors}\label{sec:detecterrors}
Finally, in order to detect translation errors, translated pairs from the previous step are checked for RTI similarity. Detecting the absensce of an RTI in a translation while avoiding false positives is non-trivial. For example, in Fig.~\ref{fig:overview}, the RTI in the first pair is ``chummy bilateral talks.'' Given the Chinese translation of the whole original sentence, it is difficult to identify which characters refer to the RTI. Words may be reordered while preserving the inherent meaning, so exact matches between RTI and container translations are not guaranteed. 

\begin{figure}[h]
\centering{}
 \includegraphics[scale=0.56]{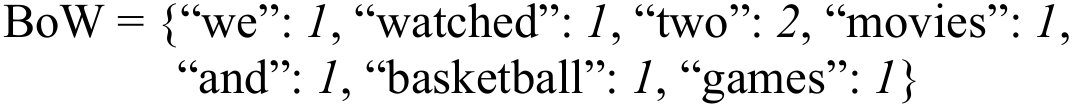} 
 \caption{Bag-of-words representation of ``we watched two movies and two basketball games.''}
 \label{fig:bagofwords}
\end{figure}

NLP techniques such as word alignment ~\cite{Liu15AAAI, Fraser07CL}, which maps a word/phrase in source text to a word/phrase in its target target, could be employed for this component of the implementation. However, performance of existing tools is poor and runtime can be quite slow. Instead, we adopt a bag-of-words (BoW) model, a representation that only considers the appearance(s) of each word in a piece of text (see Fig.~\ref{fig:bagofwords} for example). Note that this representation is a multiset. While the BoW model is simple, it has proven quite effective for modeling text in many NLP tasks. For {\implement}, using an n-gram representation of the target text provides similar performance. 

\vspace{-0.5ex}

Each translated pair consists of a translation of an RTI,  $T(r)$,\footnote{In our implementation, the context could be an empty string. Thus, $C_{empty} (r) = r$.} and of its container $T(C_{con} (r))$. After obtaining the BoWs representation of both translations ($BoW_r$ and $BoW_{con}$), the distance $dist_r(T(r), T(C_{con} (r)))$ is calculated by $dist(BoW_r, BoW_{con})$ as follows:
\begin{equation}
dist(BoW_r, BoW_{con}) = |BoW_r \setminus BoW_{con}|
\label{eq:dist}
\end{equation}
\noindent In words, this metric measures how many word occurrences are in $T(r)$ but not in $T(C_{con} (r))$. For example, the distance between ``we watch two movies and two basketball games'' ($T(C_{con} (r))$) and ``two interesting books'' ($T(r)$) is 2. If the distance is larger than a threshold $d$, which is a chosen hyperparameter, the translation pair and their source texts will be reported by our approach as a suspicious issue, indicating that at least one of the translations may contain errors. For example, in the suspicious issue in Fig.~\ref{fig:overview}, the distance is 2 because Chinese characters \begin{CJK*}{UTF8}{gbsn}亲切\end{CJK*} do not appear in the translation of the container $T(C_{con} (r))$.\footnote{For Chinese text, {\implement} regards each character as a word.}

We note that theoretically, this implementation cannot detect over-translation errors in $T(C_{con} (r))$ because additional word occurrence in $T(C_{con} (r))$ will not change the distance as calculated in Equ. \ref{eq:dist}. However, this problem does not often occur since the source text $C_{con} (r)$ is frequently the RTI in another RTI pair, in which case over-translation errors can be detected in the latter RTI pair.

\section{Evaluation}\label{sec:exper}

In this section, we evaluate the performance of {\implement} by applying it to Google Translate and Bing Microsoft Translator. Specifically, this section aims at answering the following research questions:

\begin{itemize}
  \item RQ1: How precise is the approach at finding erroneous issues?
  \item RQ2: How many erroneous translations can our approach report?
  \item RQ3: What kinds of translation errors can our approach find?
  \item RQ4: How efficient is the approach?

\end{itemize}

\subsection{Experimental Setup and Dataset}
\paragraph{Experimental environments} All experiments are run on a Linux workstation with 6 Core Intel Core i7-8700 3.2GHz Processor, 16GB DDR4 2666MHz Memory, and GeForce GTX 1070 GPU. The Linux workstation is running 64-bit Ubuntu 18.04.02 with Linux kernel 4.25.0. For sentence parsing, we use the shift-reduce parser by Zhu \emph{et al.}~\cite{SRConstParsing}, which is implemented in Stanford's CoreNLP library~\cite{stanfordcorenlp}. Our experiments consider the English$\rightarrow$Chinese language setting because of the knowledge background of the authors. 

\paragraph{Comparison} We compare {\implement} with two state-of-the-art approaches: SIT~\cite{He20ICSE} and TransRepair (ED)~\cite{Sun20ICSE}. We obtained the source code of SIT from the authors. The authors of TransRepair could not release their source code due to industrial confidentiality. Thus, we carefully implement their approach following descriptions in the paper and consulting the work's main author for crucial implementation details. TransRepair uses a threshold of 0.9 for the cosine distance of word embeddings to generate word pairs. In our experiment, we use 0.8 as the threshold because we were unable to reproduce the quantity of word pairs that the paper reported using 0.9. In this paper, we evaluate TransRepair-ED because it achieves the highest precision among the four metrics on Google Translate and better overall performance than TransRepair-LCS for Transformers (Table 2 of \cite{Sun20ICSE}). In addition, we re-tune the parameters of SIT and TransRepair using the strategies introduced in their papers. All the approaches in this evaluation are implemented in Python and released~\cite{bugsfound}.

\paragraph{Dataset} {\implement} tests machine translation software with lexically- and syntactically-correct real-world sentences. We use the dataset collected from CNN articles released by He \emph{et al.}~\cite{He20ICSE}. The details of this dataset are illustrated in Table~\ref{tab:datasets}. This dataset contains two corpora: \textit{Politics} and \textit{Business}. Sentences in the ``Politics" dataset contains 4$\sim$32 words (
the average is 19.2) and they contain 1,918 words and 933 non-repetitive words in total.  We use corpora from both categories to evaluate the performance of {\implement} on sentences with different terminology.

\begin{table}[t]
\centering{}
\caption{Statistics of input sentences for evaluation. Each corpus contains 100 sentences.}
\includegraphics[scale=0.48]{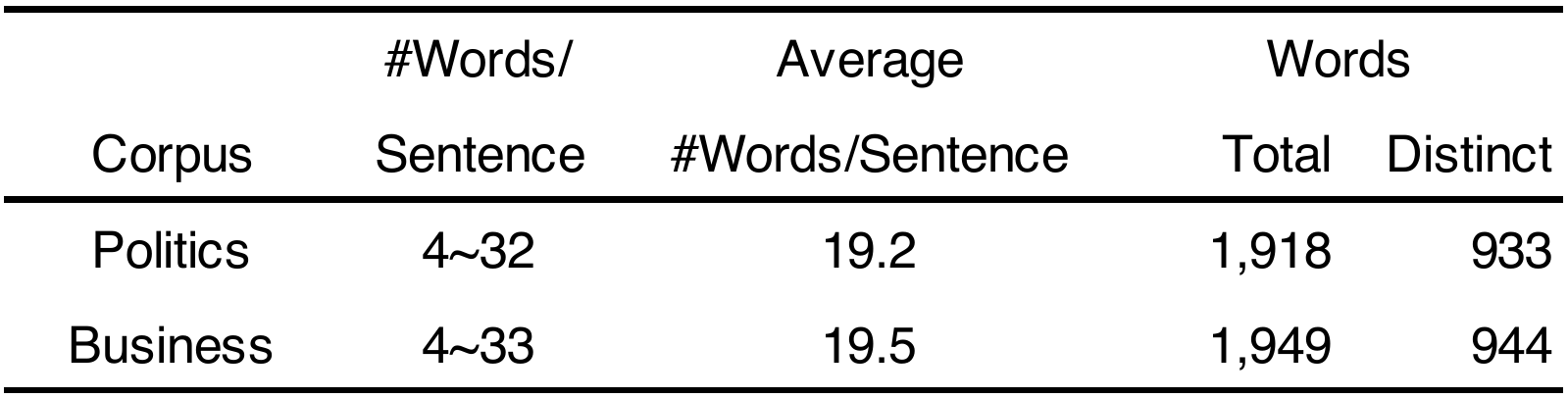}
\label{tab:datasets}
\end{table}


\subsection{Precision on Finding Erroneous Issues}\label{sec:precision}
Our approach automatically reports suspicious issues that contain inconsistent translations on the same RTI. Thus, the effectiveness of the approach lies in two aspects: (1) how precise are the reported issues; and (2) how many erroneous translations can {\implement} find? In this section, we evaluate the precision of the reported pairs, \emph{i.e.}, how many of the reported issues contain real translation errors. Specifically, we apply {\implement} to test Google Translate and Bing Microsoft Translator using the datasets characterized by Table~\ref{tab:datasets}. To verify the results, two authors manually inspect all the suspicious issues separately and then collectively decide (1) whether an issue contains translation error(s); and (2) if yes, what kind of translation error it contains.

\subsubsection{Evaluation Metric}
The output of {\implement} is a list of suspicious issues, each containing (1) an RTI, $r$, in source language and its translation, $T(r)$; and (2) a piece of text in a source language, which contains the RTI, $C_{con} (r)$, and its translation, $T(C_{con} (r))$. We define the precision as the percentage of pairs that have translation error(s) in $T(r)$ or $T(C_{con} (r))$. Explicitly, for a suspicious issue $p$, we set $error(p)$ to $true$ if $T_p (r)$ or $T_p (C_{con} (r))$ has translation error(s) (\textit{i.e.}, when the suspicious issue is an \textit{erroneous issue}). Otherwise, we set $error(p)$ to $false$. Given a list of suspicious issues, the precision is calculated by:

\begin{equation}
\label{equ:precision}
\textrm{Precision} = \frac{\sum_{p \in P} \mathds{1}\{error(p)\}}{|P|},
\end{equation}
where $P$ is the suspicious issues returned by {\implement} and $|P|$ is the number of the suspicious issues.

\subsubsection{Results}  

\begin{table*}[t]
\centering{}
\caption{{\implement}'s Precision (\# of erroneous issues/\# of suspicious issues)  using different threshold values.}
\includegraphics[scale=1]{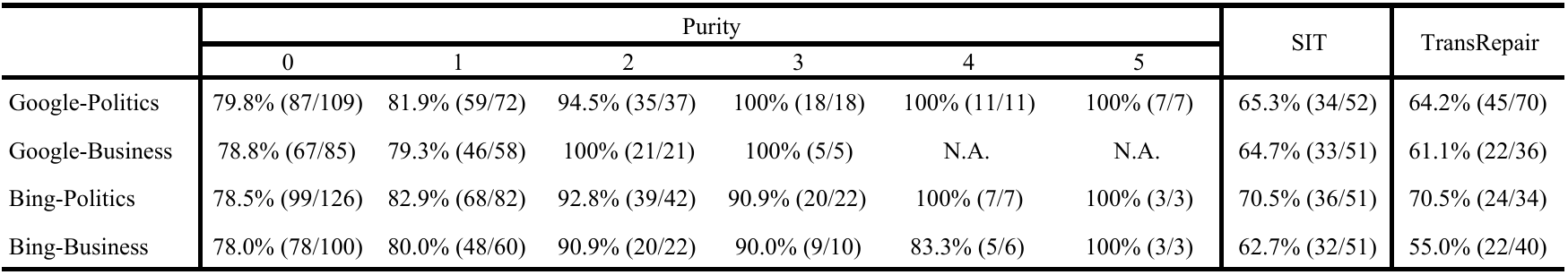}
\label{tab:precision}
\end{table*}

The results are presented in Table~\ref{tab:precision}. We observe that if the goal is to find as many issues as possible (\emph{i.e.}, $d=0$), {\implement} achieves 78\%$\sim$79.8\% precision while reporting 67$\sim$99 erroneous issues. For example, when testing Bing Microsoft Translator with the ``Business" dataset, {\implement} reports 100 suspicious issues, while 78 of them contain translation error(s), leading to 78\% precision. If we want {\implement} to be more accurate, we can use a larger distance threshold. For example, when we set the distance threshold to $5$, {\implement} achieves 100\% precision on all experimental settings. Note the precision does not increase monotonically with the threshold value. For ``Bing-Politics," the precision drops 1.9\% when changing the threshold value from $2$ to $3$. So although the number of false positives decreases, the number of true positives may decrease as well.

In our comparisons, we find {\implement} detects more erroneous issues with higher precision compared with all the existing approaches. To compare with SIT, we focus on the top-1 results (\textit{i.e.} the translation that is most likely to contain errors) reported by their system. In particular, the top-1 output of SIT contains (1) the original sentence and its translation and (2) the top-1 generated sentence and its translation. For direct comparison, we regard the top-1 output of SIT as a suspicious issue. TransRepair reports a list of suspicious sentence pairs and we regard each reported pair as a suspicious issue. Equ.~\ref{equ:precision} is used to compute the precision of the compared approaches. The results are presented in the right-most columns of Table~\ref{tab:precision}. 

When the distance threshold is at its lowest (\textit{i.e.}, $d=0$), {\implement} finds more erroneous issues with higher precision compared with SIT and TransRepair. For example, when testing Google Translate on the ``Politics" dataset, {\implement} finds 87 erroneous issues with 79.8\% precision, while SIT only finds 34 erroneous issues with 65.3\% precision. When $d=2$, {\implement} detects a similar number of erroneous issues to SIT but with significantly higher precision. For example, when testing Bing Microsoft Translator on the ``Politics" dataset, {\implement} finds 39 erroneous issues with 92.8\% precision, while SIT finds 36 erroneous issues with 70.5\% precision.\footnote{Note the precision results are different from those reported by He \emph{et al.}~\cite{He20ICSE} because Google Translate and Bing Microsoft Translator continuously update their model.} Although the precision comparison is not apples-to-apples, we believe the results have shown the superiority of {\implement}. As real-world source sentences are almost unlimited, in practice, we could set $d=2$ for this language setting to obtain a decent amount of erroneous issues with high precision.

We believe {\implement} achieves a much higher precision because of the following reasons. First, existing approaches rely on pre-trained models (\textit{i.e.}, BERT~\cite{Devlin18Bert} for SIT and GloVe~\cite{glove} and spaCy~\cite{spacy} for TransRepair) to generate sentences pairs. Although BERT should do well on this task, it could generate sentences of strange semantics, leading to false positives. Differently, {\implement} directly extract phrases from real-world sentences to construct RTI pairs and thus does not have such kind of  false positives. In addition, SIT relies on dependency parsers~\cite{Chen14EMNLP} in target sentence representation and comparison. The dependency parser could return incorrect dependency parse trees, leading to false positives.

\subsubsection{False Positives}
\begin{figure}[th]
\centering{} 
\includegraphics[scale=0.54]{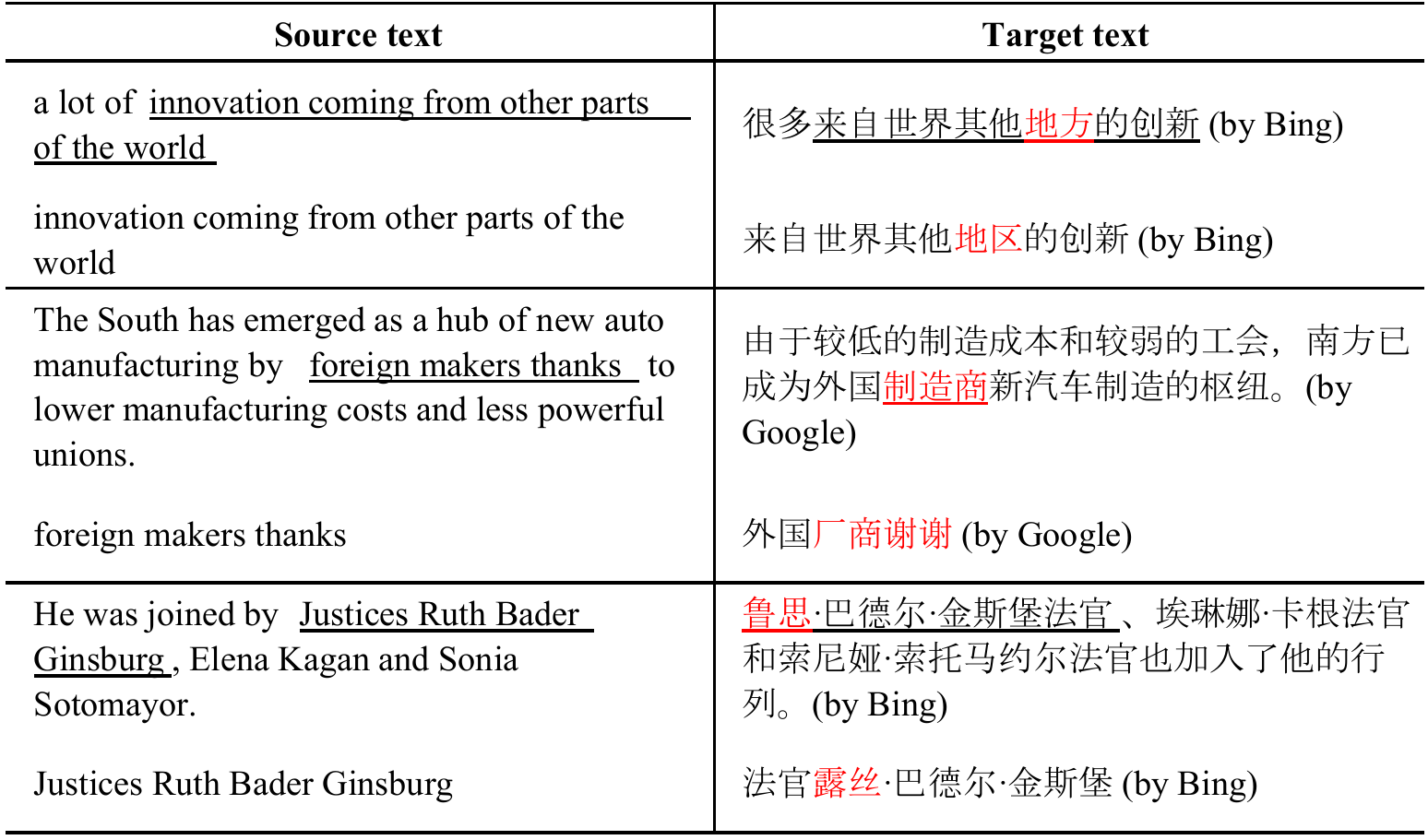}
\caption{False positive examples.}
\label{fig:falsepositive}
\end{figure}
False positives of {\implement} come from three sources. In Fig.~\ref{fig:falsepositive}, we present false positive examples when $d=0$. First, a phrase could have multiple correct translations. As shown in the first example, ``parts" have two correct translations
\begin{CJK*}{UTF8}{gbsn}(\textit{i.e.}, 地方~and 地区)\end{CJK*}
in the context ``other parts of the word". However, when $d=0$, it will be reported. This category accounts for most of {\implement}'s false positives. To alleviate this kind of false positive, we could tune the distance threshold $d$ or maintain an alternative translation dictionary. Second, the constituency parser that we use to identify noun phrases could return a non-noun phrase. In the second example, ``foreign makers thanks" is identified as a noun phrase, which leads to the change of phrase meaning. In our experiments, 6 false positives are caused by incorrect output from the constituency parser. Third, proper names are often transliterated and thus could have different correct results. In the third example, the name ``Ruth" has two correct transliterations, leading to a false positive. In our experiments 1 false positive is caused by the transliteration of proper names.

\subsubsection{RTIs Extracted by {\implement}}
We manually inspected all the 335 RTIs found by Purity. 173 RTIs were found in the ``Politics" dataset and 162 RTIs were found in the ``Business" dataset. Among these RTIs, 319 RTIs (95.2\%) should have similar translations when they are used in different contexts. The remaining 16 RTIs are caused by the errors from the constituency parser. 139 out of the 200 original sentences contain RTI(s). All the RTIs formed 620 RTI pairs. 

When the distance threshold was 2, which means the translations of the RTI could have at most two different Chinese characters, 122 RTI pairs were reported as suspicious issues, and the remaining 498 RTI pairs did not violate our assumption. As indicated in Table~\ref{tab:precision}, 115 suspicious issues are true positives, while 7 are false positives. The number of reported RTI pairs under other distance thresholds can be calculated based on the results in Table~\ref{tab:precision}.

\subsection{Erroneous Translation}
We have shown that {\implement} can report erroneous issues with high precision, where each erroneous issue contains at least one erroneous translation. Thus, to further evaluate the effectiveness of {\implement}, in this section, we study how may erroneous translations {\implement} can find. Specifically, if an erroneous translation appears in multiple erroneous issues, it will be counted once. Table~\ref{tab:recall} presents the number of erroneous translations under the same experimental settings as in Table~\ref{tab:precision}. We can observe that when $d=0$, {\implement} found 54$\sim$74 erroneous translations. If we intend to have a higher precision by setting a larger distance threshold, we will reasonably obtain fewer erroneous translations. For example, if we want to achieve 100\% precision, we can obtain 32 erroneous translations in Google Translate ($d=3$).  
\begin{table}[t]
\centering{}
\caption{The number of translations that contain errors using different threshold values.}
\includegraphics[scale=0.68]{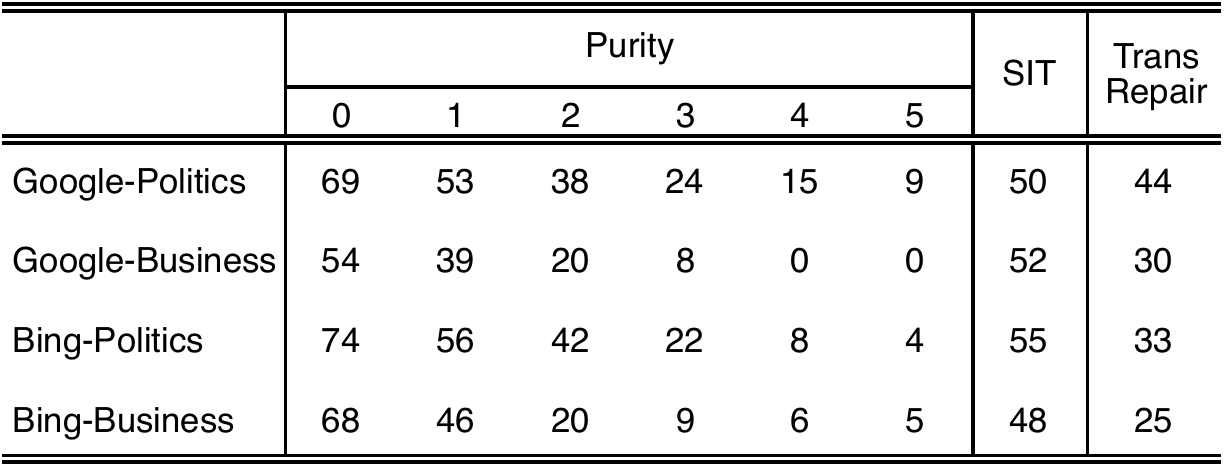}
\label{tab:recall}
\end{table}

\begin{figure}[t]
\centering{}
 \includegraphics[scale=0.29]{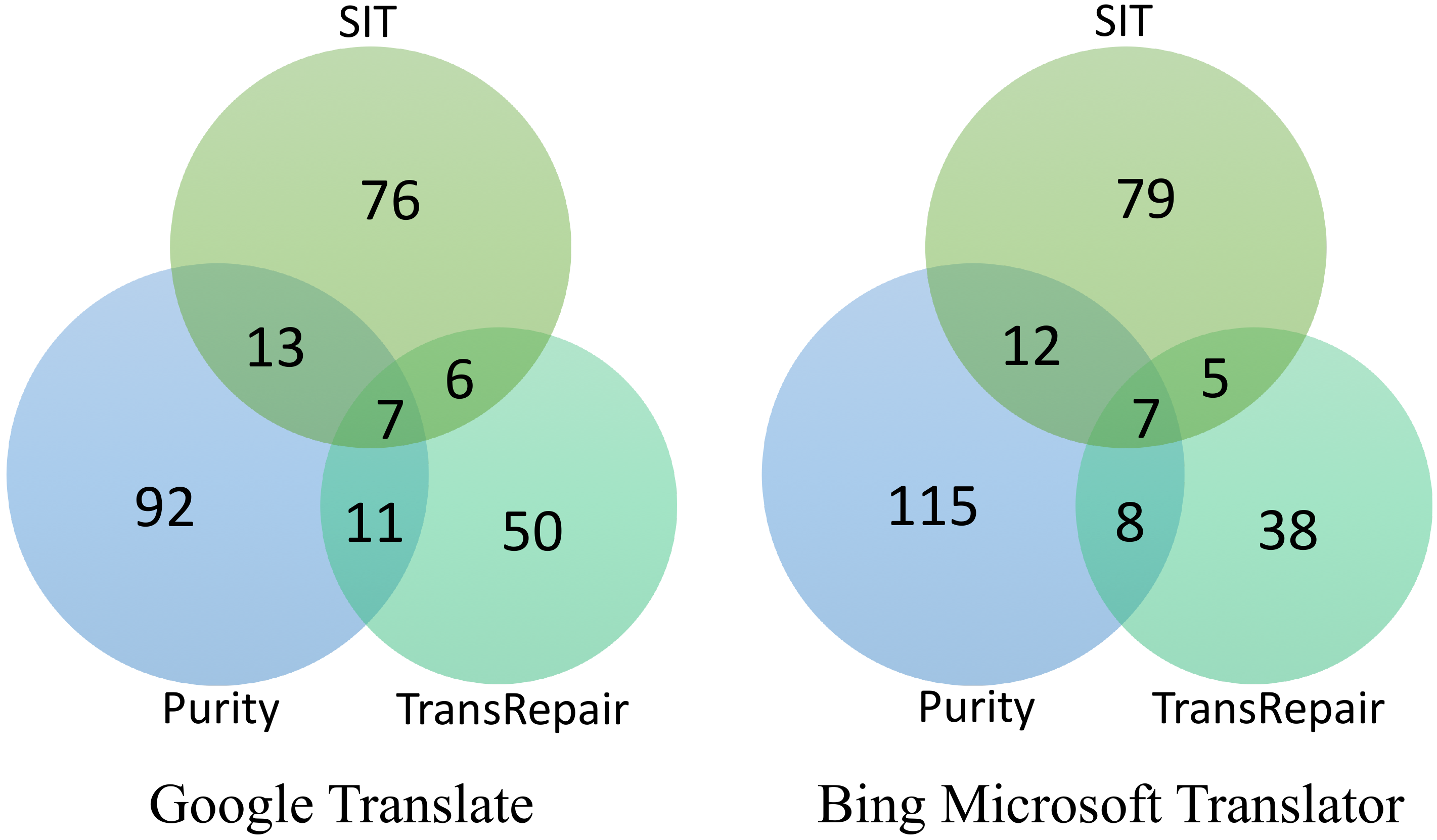} 
\caption{Erroneous translations reported by {\implement}, SIT,  and TransRepair.}
\label{fig:venn}
\end{figure}

We further study the erroneous translations found by {\implement}, SIT and TransRepair. Fig.~\ref{fig:venn} demonstrates the results via Venn diagrams. We can observe that, 7 erroneous translations from Google Translate and 7 erroneous translations from Bing Microsoft Translator can be detected by all the three approaches. These are the translations for some of the original source sentences. 207 erroneous translations are unique to {\implement} while 155 erroneous translations are unique to SIT and 88 erroneous translations are unique to TransRepair. After inspecting all the erroneous translations, we find that {\implement} is effective at reporting translation errors for phrases. Meanwhile, the unique errors to SIT are mainly from similar sentence of one noun or adjective difference. The unique errors to TransRepair mainly come from similar sentences of one number difference (\textit{e.g.}, ``five" $\rightarrow$ ``six"). Based on these results, we believe our approach complements the state-of-the-art approaches.

\subsection{Types of Reported Translation Errors}
{\implement} is capable of detecting translation errors of diverse kinds. Specifically, in our evaluation, {\implement} has successfully detected 5 kinds of translation errors: under-translation, over-translation, word/phrase mistranslation, incorrect modification, and unclear logic. Table~\ref{tab:bugtypes} presents the number of translations that have a specific kind of error. We can observe that word/phrase mistranslation and unclear logic are the most common translation errors.

To provide a glimpse of the diversity of the uncovered errors, this section highlights examples of all the 5 kinds of errors. The variety of the detected translation errors demonstrates RTI's (offered by {\implement}) efficacy and broad applicability. We align the definition of these errors with SIT~\cite{He20ICSE} because it is the first work that found and reported these 5 kinds of translation errors.

\begin{table}[t]
\centering{}
\caption{Number of translations that have specific errors in each category.}
\includegraphics[scale=0.5]{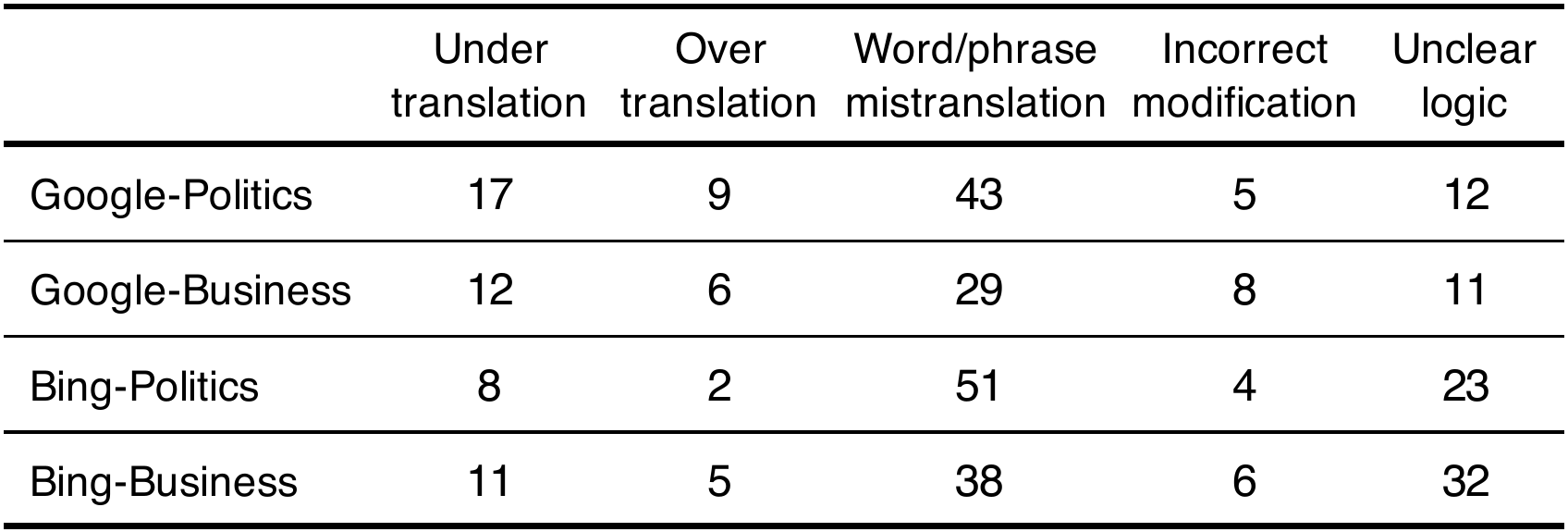}
\label{tab:bugtypes}
\end{table}

\subsubsection{Under-translation}
\begin{figure}[h]
\centering{} 
\includegraphics[scale=0.6]{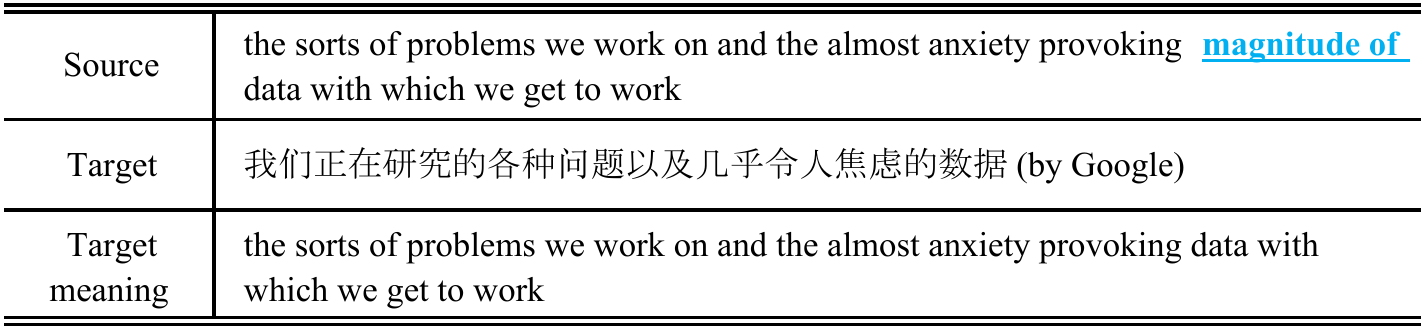}
\caption{Example of under-translation error detected.}
\label{fig:undertranslation}
\end{figure}

If some parts of the source text are not translated in the target text, it is an under-translation error. For example, in Fig.~\ref{fig:undertranslation}, ``magnitude of" is not translated by Google Translate. Under-translation often leads to target sentences of different semantic meanings and the lack of crucial information. Fig.~\ref{fig:overview} also reveals an under-translation error. In this example, the source text emphasizes that the bilateral talks are chummy while this key information is missing in the target text.

\subsubsection{Over-translation}
\begin{figure}[h]
\centering{} 
\includegraphics[scale=0.6]{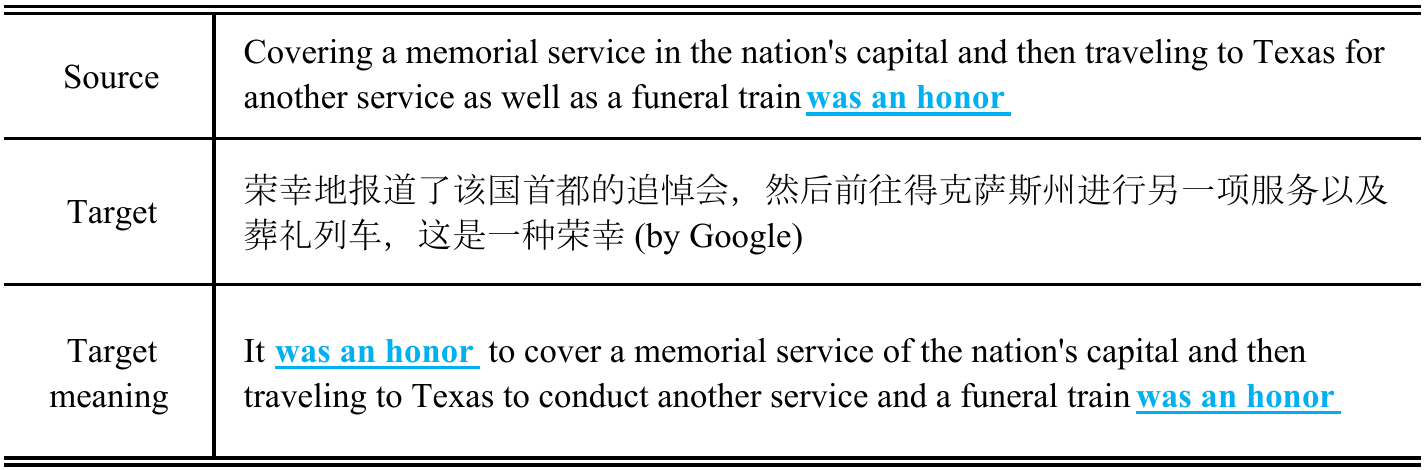}
\caption{Example of over-translation error detected.}
\label{fig:overtranslation}
\end{figure}

If some parts of the target text are not translated from word(s) of the source text or some parts of the source text are unnecessarily translated for multiple times, it is an over-translation error. In Fig.~\ref{fig:overtranslation}, ``was an honor" is translated twice by Google Translate in the target text while it only appears once in the source text, so it is an over-translation error. Over-translation brings unnecessary information and thus can easily cause misunderstanding. 

\subsubsection{Word/phrase Mistranslation}
\begin{figure}[h]
\centering{} 
\includegraphics[scale=0.6]{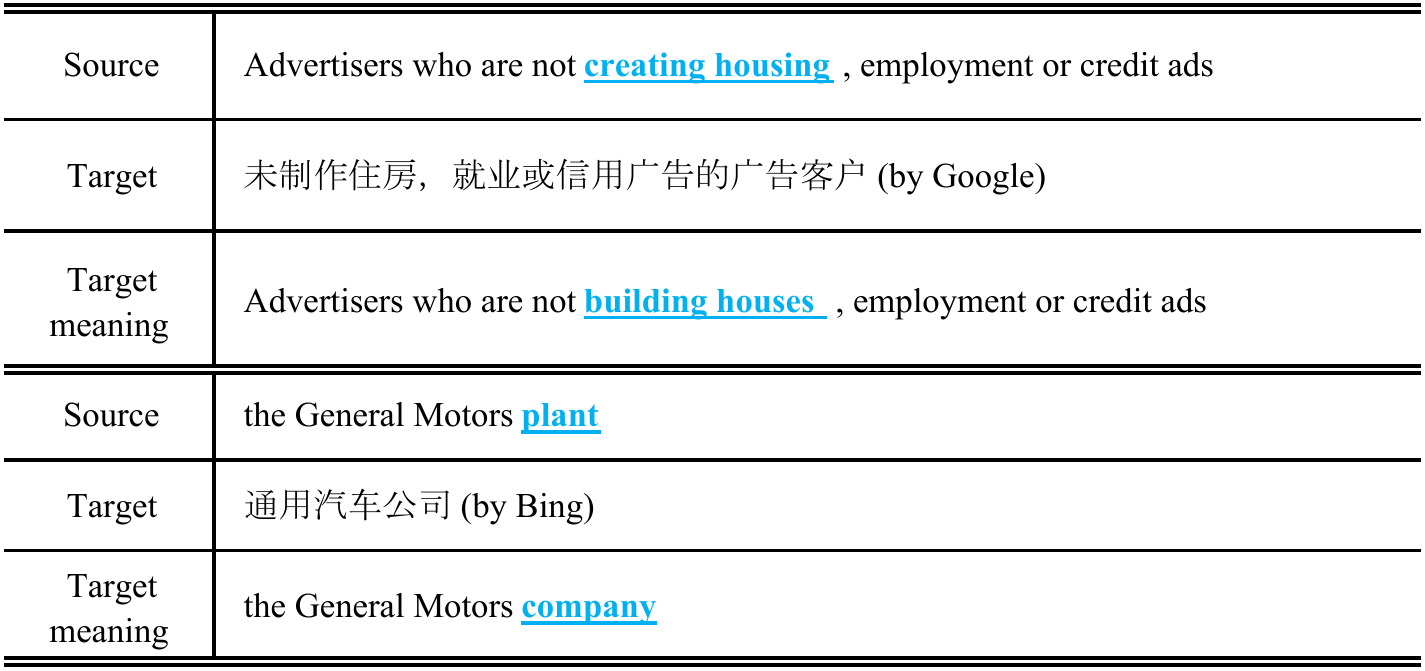}
\caption{Examples of word/phrase mistranslation errors detected.}
\label{fig:wordphrase}
\end{figure}

If some words or phrases in the source text is incorrectly translated in the target text, it is a word/phrase mistranslation error. In Fig.~\ref{fig:wordphrase}, ``creating housing" is translated to ``building houses" in the target text. This error is caused by ambiguity of polysemy. The word ``housing" means ``a general place for people to live in" or ``a concrete building consisting of a ground floor and upper storeys." In this example, the translator mistakenly thought ``housing" refers to the later meaning, leading to the translation error. In addition to ambiguity of polysemy, word/phrase mistranslation can be also caused by the surrounding semantics. In the second example of Fig.~\ref{fig:wordphrase}, ``plant" is translated to ``company" in the target text. We think that in the training data of the NMT model, ``General Motors" often has the translation ``General Motors company", which leads to a word/phrase mistranslation error in this scenario.

\subsubsection{Incorrect Modification}
\begin{figure}[h]
\centering{} 
\includegraphics[scale=0.6]{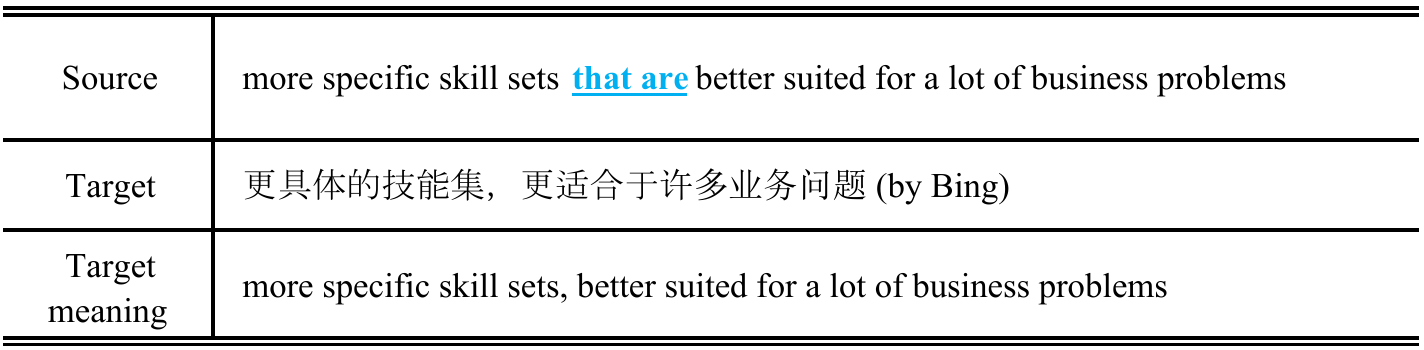}
\caption{Example of incorrect modification error detected.}
\label{fig:modification}
\end{figure}

If some modifiers modify the wrong element, it is an incorrect modification error. In Fig.~\ref{fig:modification}, ``better suited for a lot of business problems" should modify ``more specific skill sets". However, Bing Microsoft Translator inferred they are two separate clauses, leading to an incorrent modification error.

\subsubsection{Unclear Logic}
\begin{figure}[h]
\centering{} 
\includegraphics[scale=0.6]{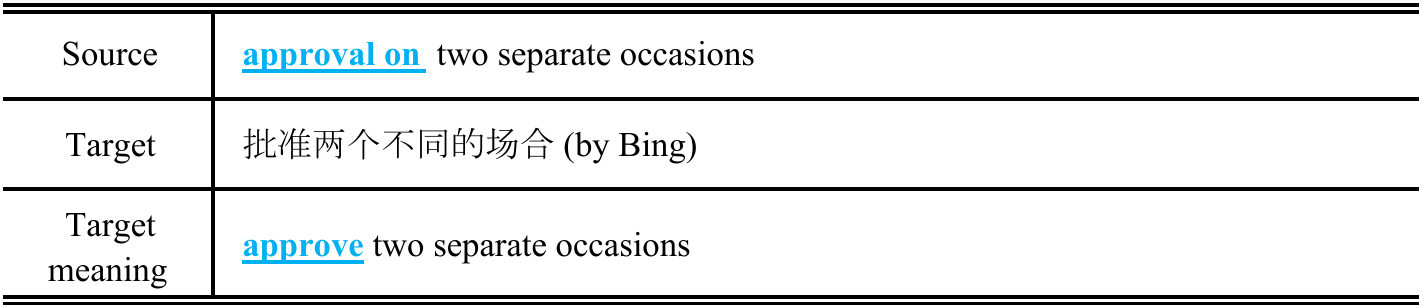}
\caption{Example of unclear logic error detected.}
\label{fig:logic}
\end{figure}

If all the words are correctly translated but the logic of the target text is wrong, it is an unclear logic error. In Fig.~\ref{fig:logic}, Bing Microsoft Translator correctly translated ``approval" and ``two separate occasions". However, Bing Microsoft Translator returned ``approve two separate occasions" instead of ``approval on two separate occasions" because the translator does not understand the logical relation between them. Fig.~\ref{fig:insight} also demonstrates an unclear logic error. Unclear logic errors widely exist in the translations returned by modern machine translation software, which is to some extent a sign of whether the translator truely understands certain semantic meanings.

\subsection{Running Time}
In this section, we study the efficiency (\textit{i.e.}, running time) of {\implement}. Specifically, we adopt {\implement} to test Google Translate and Bing Microsoft Translator with the ``Politics" and the ``Business" dataset. For each experimental setting, we run {\implement} 10 times and use the average time as the final result. Table~\ref{tab:efficiency} presents the total running time of {\implement} as well as the detailed running time for initialization, RTI pairs construction, translation collection, and referential transparency violation detection.
\begin{table}[t]
\centering{}
\caption{Running time of {\implement} (sec)}
\includegraphics[scale=0.75]{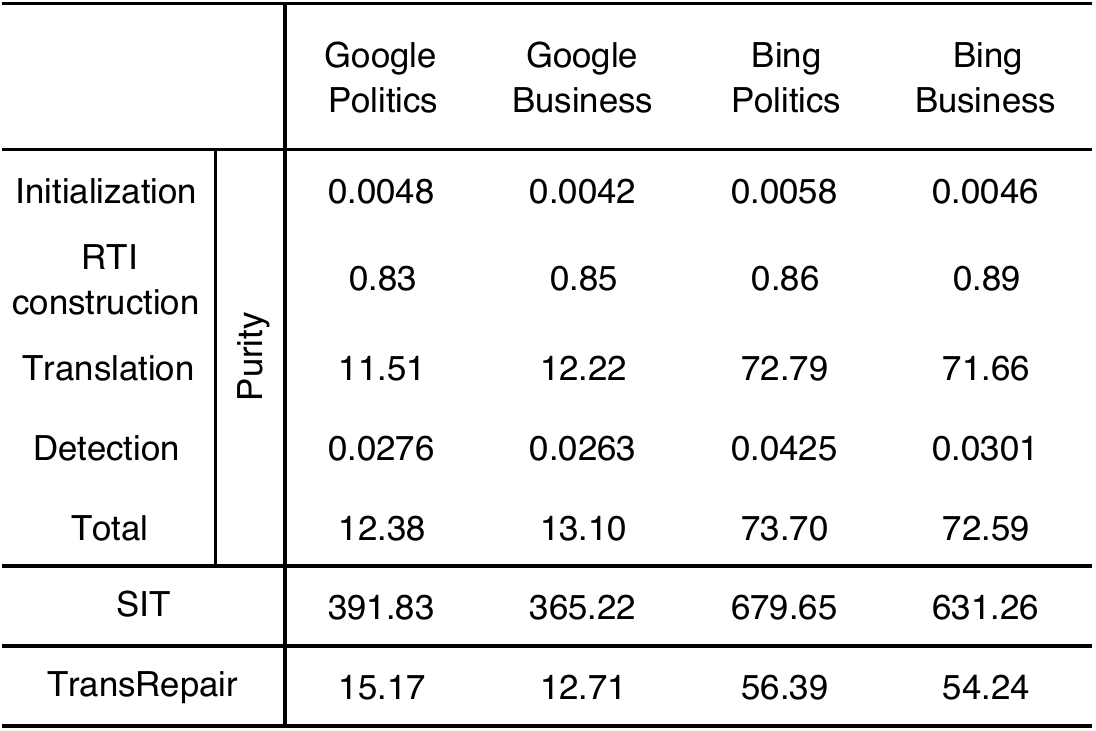}
\label{tab:efficiency}
\end{table}

We can observe that {\implement} spent less than 15 seconds on testing Google Translate and around 1 minute on testing Bing Microsoft Translator. Specifically, more than 90\% of the time is used in the collection of translations via translators' APIs . In our implementation, we invoke the translator API once for each piece of source text and thus the network communication time is included. If developers intend to test their own machine translation software with {\implement}, the running time of this step will be even less.

Table~\ref{tab:efficiency} also presents the running time of SIT and TransRepair using the same experimental settings. SIT spent more than 6 minutes to test Google Translate and around 11 minutes to test Bing Microsoft Translator. This is mainly because SIT translates 44,414  words for the ``Politics" dataset and 41,897 words for the ``Business" dataset. Meanwhile, {\implement} and TransRepair require fewer translations (7,565 and 6,479 for {\implement} and 4,271 and 4,087 for TransRepair). Based on these results, we conclude that {\implement} achieves comparable efficiency to the state-of-the-art methods.

\subsection{Fine-tuning with Errors Reported by {\implement}}\label{sec:finetune}
The ultimate goal of testing is to improve software robustness. Thus, in this section, we study whether reported mistranslations can act as a fine-tuning set to both improve the robustness of NMT models and quickly fix errors found during testing. Fine-tuning is a common practice in NMT since the domain of the target data (i.e. data used at runtime) is often different than that of the training data ~\cite{Sennrich15ACL, Chu17ACL}. To simulate this situation, we train a transformer network with global attention \cite{Vaswani17NeurIPS}---a standard architecture for NMT models---on the WMT'18 ZH--EN (Chinese-to-English) corpus \cite{bojar-EtAl:2018:WMT1}, which contains $\sim$ 20M sentence pairs. We reverse the standard direction of translation (\textit{i.e.} to EN--ZH) for comparison with our other experiments. We use the fairseq framework \cite{fairseq} to create the model. 

To test our NMT model, we crawled the 10 latest articles under the ``Entertainment'' category of CNN website and randomly extract 80 English sentences. The dataset collection process aligns with that of the ``Politics'' and the ``Business'' datasets \cite{He20ICSE} used in the main experiments. We run {\implement} with the ``Entertainment" dataset using our trained model as the system under test; {\implement} successfully finds 42 erroneous translations. We manually label them with correct translations and fine-tune the NMT model on these 42 translation pairs for 8 epochs---until loss on the WMT'18 validation set stops decreasing. After this fine-tuning, 40 of the 42 sentences are correctly translated. One of the two translations that were not corrected can be attributed to parsing errors; while the other (source text: ``one for Best Director") has an ``ambiguous reference" issue, which essentially makes it difficult to translate without context. Meanwhile, the BLEU score on the WMT'18 validation set stayed well within standard deviation \cite{post2018clarity}. This demonstrates that error reported by {\implement} can indeed be fixed without retraining a model from scratch -- a resource and time intensive process.

\section{Discussion}\label{sec:dis}

\subsection{RTI for Robust Machine Translation}
In this section, we discuss the utility of referential transparency towards building robust machine translation software. Compared with traditional software, the error fixing process of machine translation software is arguably more difficult because the logic of NMT models lies within a complex model structure and its parameters rather than human-readable code. Even if the computation which causes a mistranslation can be identified, it is often not clear how to change the model to correct the mistake without introducing new errors. While model correction is a difficult open problem and is not the main focus of our paper, we find it important to explain that the translation errors found by {\implement} can be used to both fix and improve machine translation software.

For online translation systems, the fastest way to fix a mistranslation is to hard-code the translation pair. Thus, the translation errors found by {\implement} can be quickly and easily addressed by developers to avoid mistranslations that may lead to negative effects~\cite{translation1, translation2, translation3, translation4, translation5, translation6}. The more robust solution is to incorporate the mistranslation into the training dataset. In this case, a developer can add the source sentence of a translation error along with its correct translation to the training set of the neural network and retrain or fine-tune the network. While retraining a large neural network from scratch can take days, fine-tuning on a few hundred mistranslations takes only a few minutes, even for the large, SOTA models. We note that this method does not absolutely guarantee the mistranslation will be fixed, but our experiments (Section~\ref{sec:finetune}) show it to be quite effective in resolving errors. The developers may also find the reported issues useful for further analysis/debugging because it resembles debugging traditional software via input minimization/localization. In addition, as RTI's reported results are in pairs, they can be utilized as a dataset for 
future empirical studies on translation errors.

\subsection{Change of Language}
In our implementation, {\implement}, we use English as the source language and Chinese as the target language. To match our exact implementation, there needs to be a constituency parser---or data to train such a parser---available in the chosen source language, as this is how we find RTIs. The Stanford Parser\footnote{https://nlp.stanford.edu/software/lex-parser.html\#Download} currently supports six languages. Alternatively, one can train a parser following, for example, Zhu \textit{et al}.~\cite{SRConstParsing}. Other modules of {\implement} remain unchanged. Thus, in principle, it is quite easy to re-target RTI to other languages. Note that while we expect the RTI property to hold for most of the languages, there may be confounding factors in the structure of a language that break our assumptions.

\section{Related Work}\label{sec:related}

\subsection{Robustness of AI Software}
Recently, Artificial Intelligence (AI) software has been adopted by many domains; this is largely due to the modelling abilities of deep neural networks. However, these systems can generate erroneous outputs that, e.g., lead to fatal accidents~\cite{accident1, accident2, accident3}. To explore the robustness of AI software, a line of research has focused on attacking different systems that use deep neural networks, such as autonomous cars~\cite{Goodfellow15ICLR, Athalye18ICML} and speech recognition services~\cite{Carlini16Security, Du19Arxiv}. These work aim to fool AI software by feeding input with imperceptible perturbations (\textit{i.e.}, adversarial examples). Meanwhile, researchers have also designed approaches to improve AI software's robustness, such as robust training mechanisms~\cite{Madry18ICLR, Lin19ICLR, Mao19NeurIPS}, adversarial examples detection approaches~\cite{Tao18NeurIPS, Wang19ICSE}, and testing/debugging techniques~\cite{Pei17SOSP, Tian18ICSE, Ma18ASE, Ma18FSE, Kim19ICSE, JZhang19Arxiv, Du19FSE, Xie19ISSTA}. Our paper also studies the robustness of a widely-adopted AI software, but focuses on machine translation systems, which has not been explored by these papers. Additionally, most of these approaches are white-box, utilizing gradients/activation values, while our approach is black-box, requiring no model internal details at all.

\subsection{Robustness of NLP Systems}
Inspired by robustness studies in the computer vision field, NLP (natural language processing) researchers have started exploring attack and defense techniques for various NLP systems. Typical examples include sentiment analysis~\cite{Iyyer18NAACL, Alzantot18EMNLP, Li19NDSS, Pruthi19ACL, Ribeiro18ACL}, textual entailment~\cite{Iyyer18NAACL}, and toxic content detection~\cite{Li19NDSS}. However, these are all basic classification tasks while machine translation software is more complex in terms of both model output and network structure.

The robustness of other complex NLP systems has also been studied in recent years. Jia and Liang~\cite{Jia17EMNLP} proposed a robustness evaluation scheme for the Stanford Question Answering Dataset (SQuAD), which is widely used in the evaluation of reading comprehension systems. They found that even the state-of-the-art system, achieving near human-level F1-score, fails to answer questions about paragraphs correctly when an adversarial sentence is inserted. Mudrakarta \emph{et al.}~\cite{Mudrakarta18ACL} also generate adversarial examples for question answering tasks on images, tables, and passages of text. These approaches typically perturb the system input and assume that the output (\textit{e.g.}, a person name or a particular year) should remain the same. However, the output of machine translation (\textit{i.e.}, a piece of text) is more complex. In particular, one source sentence could have multiple correct target sentences. Thus, testing machine translation software, which is the goal of this paper, is more difficult.

\subsection{Robustness of Machine Translation}
Recently, researchers have started to explore the robustness of NMT models. Belinkov and Bisk~\cite{Belinkov18ICLR} found that both synthetic (\textit{e.g.}, character swaps) and natural (\textit{e.g.}, misspellings) noise in source sentences could break character-based NMT models. In contrast, our approach aims to find lexically- and syntactically-correct source texts that lead to erroneous output by machine translation software, which the errors more commonly found in practice. To improve the robustness of NMT models, various robust training mechanisms have been studied~\cite{Cheng18ACL, Cheng19ACL}. In particular, noise is added to the input and/or the internal network embeddings during training. Different from these approaches, we focus on testing machine translation.

Zheng \emph{et al.}~\cite{Zheng18Arxiv} proposed specialized approaches to detect under- and over-translation errors respectively. Different from them, our approach aims at finding general errors in translation. He \textit{et al}.~\cite{He20ICSE} and Sun \textit{et al}.~\cite{Sun20ICSE} proposed metamorphic testing methods for general translation errors: they compare the translations of two similar sentences (\textit{i.e.}, differed by one word) by sentence structures ~\cite{He20ICSE} and four existing metrics on sub-strings~\cite{Sun20ICSE} respectively. In addition, Sun \textit{et al}.~\cite{Sun20ICSE} designed an automated translation error repair mechanism. Compared with these approaches, RTI can find more erroneous translations with higher precision and comparable efficiency. The translation errors reported are diverse and distinguished from those found by existing papers~\cite{He20ICSE, Sun20ICSE}. Thus, we believe RTI can compliment with the state-of-the-art approaches. Gupta \emph{et al.}~\cite{Gupta20FSE} developed a translation testing approach based on pathological invariance: sentences of different meanings should not have identical translation. We did not compare with this paper because it is based on an orthogonal approach and we consider it as a concurrent work.

\subsection{Metamorphic Testing}
The key idea of metamorphic testing is to detect violations of metamorphic relations across input-output pairs. Metamorphic testing has been widely employed to test traditional software, such as compilers~\cite{Le14PLDI, Lidbury15PLDI}, scientific libraries~\cite{Zhang14ASE, Kanewala16STVR}, and service-oriented applications~\cite{Chan05QSIC_B, Chan07IJWSR}. Because of its effectiveness on testing ``non-testable" systems, researchers have also designed metamorphic testing techniques for a variety of AI software. Typical examples include autonomous cars~\cite{Tian18ICSE, Zhang18ASE}, statistical classifiers~\cite{Xie09QSIC, Xie11JSS}, and search engines~\cite{Zhou16TSE}. In this paper, we introduce a novel metamorphic testing approach for machine translation software.

\section{Conclusion}\label{sec:con}

We have presented a general concept---referentially transparent input (RTI)---for testing machine translation software. In contrast to existing approaches, which perturb a word in natural sentences (\textit{i.e.}, the context is fixed) and assume that the translation should have only small changes, this paper assumes the RTIs should have similar translations across different contexts. As a result, RTI can report different translation errors (\textit{e.g.}, errors in the translations of phrases) of diverse kinds and thus complements existing approaches. The distinctive benefits of RTI are its simplicity and wide applicability. We have used it to test Google Translate and Bing Microsoft Translator and  found 123 and 142 erroneous translations respectively with the state-of-the-art running time, clearly demonstrating the ability of RTI---offered by {\implement}---to test machine translation software. For future work, we will continue refining the general approach and extend it to other RTI implementations, such as using verb phrases as RTIs or regarding whole sentences as RTIs, pairing them with the concatenation of a semantically-unrelated sentence. We will also launch an extensive effort on translation error diagnosis and automatic repair for machine translation systems.

\section*{Acknowledgments}
We thank the anonymous ICSE reviewers for their valuable feedback on the earlier draft of this paper. In addition, the tool implementation benefited tremendously from Stanford NLP Group's language parsers~\cite{stanfordcorenlp}.

\balance
\bibliographystyle{IEEEtran}
\bibliography{References}

\end{document}